\documentclass[10pt,twocolumn,letterpaper]{article}

\usepackage{iccv}
\usepackage{times}
\usepackage{epsfig}
\usepackage{graphicx}
\usepackage{amsmath}
\usepackage{amssymb}

\usepackage{caption}
\usepackage{subcaption}
\usepackage{array,multirow}
\usepackage{threeparttable}
\usepackage{booktabs}

\usepackage[pagebackref=true,breaklinks=true,letterpaper=true,colorlinks,bookmarks=false]{hyperref}

\iccvfinalcopy 

\ificcvfinal\fi

\begin{document}

\title{Remote Pulse Estimation in the Presence of Face Masks}

\author{Jeremy Speth, Nathan Vance, Patrick Flynn, Kevin Bowyer, Adam Czajka\\
University of Notre Dame\\
{\tt\small \{jspeth, nvance1, flynn, kwb, aczajka\}@nd.edu}
}

\maketitle
\ificcvfinal\thispagestyle{empty}\fi

\maketitle

\begin{abstract}
   Remote photoplethysmography (rPPG), a family of techniques for monitoring blood volume changes, may be especially useful for widespread contactless health monitoring 
   using face video from consumer-grade visible-light cameras. The COVID-19 pandemic has caused the widespread use of protective face masks. We found that occlusions from cloth face masks increased the mean absolute error of heart rate estimation by more than 80\% when deploying methods designed on unmasked faces. We show that augmenting unmasked face videos by adding patterned synthetic face masks forces the model to attend to the periocular and forehead regions, improving performance and closing the gap between masked and unmasked pulse estimation. To our knowledge, this paper is the first to analyse the impact of face masks on the accuracy of pulse estimation and offers several novel contributions: (a) 3D CNN-based method designed for remote photoplethysmography in a presence of face masks, (b) two publicly available pulse estimation datasets acquired from 86 unmasked and 61 masked subjects, (c) evaluations of handcrafted algorithms and a 3D CNN trained on videos of unmasked faces and with masks synthetically added, and (d) data augmentation method to add a synthetic mask to a face video.
\end{abstract}

\section{Introduction}

Remote pulse detection is especially useful in settings where health diagnostics are desired, but using contact sensors is infeasible, presents some risk, or professional sensors (\eg pulse oximeters) are not available. The COVID-19 pandemic is one such scenario where extracting cardiac diagnostics without surface contact mitigates the risk of viral transmission, and can potentially allow for ubiquitous health monitoring at a critical time. 

\begin{figure*}
    \centering
    \includegraphics[width=\linewidth]{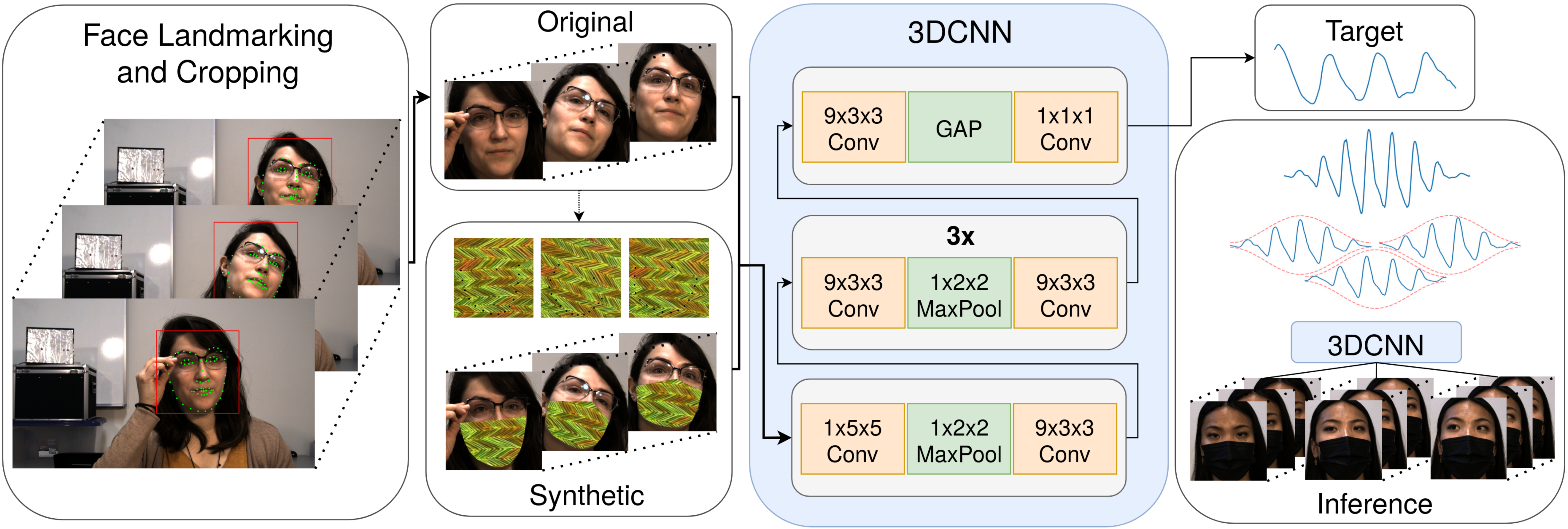}
    \caption{Training and inference pipeline for the spatiotemporal modeling task of remote pulse estimation. Raw RGB frames are landmarked and cropped, and  then either fed directly into the 3DCNN or a synthetic mask is added. Multiple frame sequences are overlap-added \cite{DeHaan2013} to produce the full pulse waveform.}
    \label{fig:main_figure}
\end{figure*}

The widespread adoption of face mask usage has caused significant problems for existing technologies that assume an unobstructed view of the face \cite{NIST2020}. Nearly all recent rPPG algorithms extract the signal from the face \cite{Chen2018, DeHaan2013, Poh2011, Wang2017, Yu2019}, sometimes even limiting the analyzed region to the cheeks \cite{Li2014, Tulyakov2016}, which of course is a region of the face generally occluded by a mask. This methodological choice presents a risk to the aforementioned algorithms due to a smaller spatial region for increased risk of signal contamination by noise. While early contactless pulse estimation algorithms used hand-crafted features in both the temporal and spatial domains, more recent works have shown that convolutional neural networks (CNN) fed with spatiotemporal representations may outperform handcrafted approaches \cite{Chen2018, Yu2019, Niu2020}. We thus select a 3DCNN as the model architecture for this work, shown in Fig. \ref{fig:main_figure}.

To accommodate the research community's need for large-scale realistic physiological datasets, we present two new datasets. The first collection was recorded from 86 unmasked subjects in an interview setting with natural conversation. This dataset is used for designing and fine-tuning all the methods considered in this work. The second database, collected in the same operational setting from 61 masked subjects, is designed for subject-disjoint evaluation of the effects of masks on estimation of pulse waveforms.

The {\bf main novel contributions of this work} are (1) the first face video dataset of masked individuals for remote physiological monitoring, and (2) the first baseline experiments and systematic analysis of the effects of face masks on remote pulse estimation performance, {\bf answering three research questions:}
\begin{enumerate}
\setlength{\itemsep}{0pt}
        \item[(Q1)] Is the accurate pulse rate estimation possible on subjects wearing masks?
        \item[(Q2)] If the answer to (Q1) is affirmative, does inclusion of face videos with synthetic masks result in better performance on videos of subjects wearing actual masks?
        \item[(Q3)] What adaptations to the existing rPPG methods are useful to fine-tune them to COVID-19 and future health crises?
\end{enumerate}

To the authors' knowledge, this is the first paper to explore the effects of face masks on the accuracy of remote pulse detection algorithms.
\section{Background and Related Work}\label{sec:background}

Remote photoplethysmography is the process of estimating the blood volume pulse (BVP) by observing changes in the light reflected from the skin. Microvasculature beneath the skin's surface fills with blood, which changes reflected color due to the light absorption of hemoglobin. In practice, color changes attributable to blood volume are subtle and may be obscured by variations due to factors such as illumination changes and body movements. The problem is further compounded by face masks, which decrease surface area available to detect the pulse, and consequently, decrease sample size for signal extraction.

Due to the difficulty of extracting the pulse signal from the optical signal, early studies began with stationary subjects and manually selected skin regions  \cite{Wieringa2005, Verkruysse2008}. An early advance was developed by Poh \etal \cite{Poh2010, Poh2011} when they applied blind source separation through independent component analysis (ICA) to the red, green, and blue color channels. Several advancements combined color channels in meaningful ways to locate the pulse signal \cite{DeHaan2013, DeHaan2014, Wang2016, Wang2017}. The first approach (CHROM) considered the chrominance signal, which was agnostic to illumination changes and robust to movement \cite{DeHaan2013}. Later improvements relaxed assumptions on the distortion signals from movement \cite{DeHaan2014}, and examined rotation of the skin pixels' subspace \cite{Wang2016}. Lastly, \cite{Wang2017} introduced the plane orthogonal to skin (POS) algorithm, which defines and updates a projection direction for separating the specular and pulse components.

Until Li \etal \cite{Li2014} designed an effective pulse detector on the MAHNOB-HCI database \cite{Soleymani2012}, many approaches had been designed and tested on relatively small private datasets. After using the public MAHNOB-HCI dataset many groups were able to compare their estimators \cite{Li2014, Tulyakov2016, Chen2018, Yu2019}, and it spurred the creation of more public datasets such as MMSE-HR \cite{Tulyakov2016} and VIPL-HR \cite{Niu2018}. The increased size of datasets made it possible to train deep neural networks. The first deep learning approach \cite{Hsu2014} trained a regression model on ICA and chrominance features.

Later, deep learning models for rPPG were trained on the spatial \cite{Hsu2017, Chen2018, Niu2020} and spatiotemporal \cite{Yu2019} dimensions of the video rather than extracted temporal features alone. Hsu \etal \cite{Hsu2017} trained VGG-15 on images of the frequency representation of the averaged color signal to predict heart rate. Chen \etal \cite{Chen2018} took inspiration from two-stream networks \cite{Simonyan2014} and simultaneously fed frame differences and raw frames to a two-stream CNN, predicting the derivative of the waveform. A recent approach extracted spatial-temporal maps from a grid of regions over the face, and fed each averaged region into ResNet-18 followed by a single gated recurrent unit (GRU) to predict single heart rate \cite{Niu2020}.

Yu \etal \cite{Yu2019} constructed a 3DCNN which was given video clips and minimized the negative Pearson correlation between the ground truth waveform and their output waveform. The main advantage of their spatiotemporal network over the networks in \cite{Hsu2014, Hsu2017, Niu2020}, is its capability of producing a waveform, rather than a single value for the signal's frequency. They predict several cardiac metrics such as the respiratory frequency and heart rate variability. PPG waveforms have also shown to be useful for predicting blood pressure \cite{Martinez2018}. Due to this advantage, {\bf we design a similar 3DCNN architecture with modifications to the temporal dimensions of the spatiotemporal kernels}, such that longer-range time dependencies can be captured.

While rPPG has been used for many applications such as presentation attack detection \cite{Heusch2018, Liu2018} to distinguish between no pulse detected (presentation attack) and pulse detection (live), our goal is to determine how accurately the pulse rate can be estimated from a known live face wearing a mask. {\bf Face occlusions have never been explored in rPPG}.
\section{Datasets}

\begin{figure*}
    \centering
    \includegraphics[width=\linewidth]{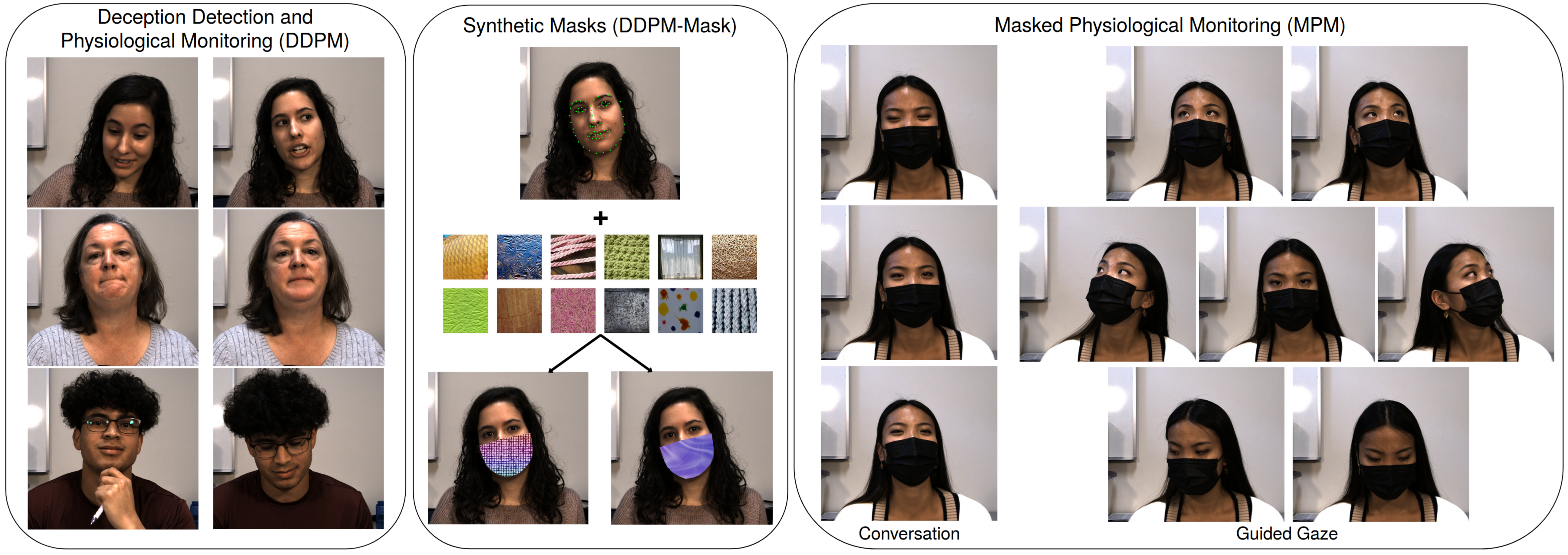}
    \caption{Frames from all pulse databases used throughout this paper are shown for the unmasked, synthetically masked, and masked videos. Patterns for the synthetic masks are randomly sampled from the Describable Textures Dataset \cite{Cimpoi14}.}
    \label{fig:masked_db}
\end{figure*}

We present two new datasets for remote physiological monitoring. The first was recorded prior to the COVID-19 outbreak, so all subjects are unmasked. The second was recently collected to assess remote pulse detection algorithms in the presence of face masks. Both datasets were collected from consenting subjects under a human subjects research protocol approved by the authors' Human Subjects Institutional Review Board. Both datasets were recorded in the same setting with subjects seated approximately 1 to 2 meters from the RGB camera. The ground truth heart rate, blood oxygenation, and blood volume pulse waveforms were collected by a Contec CMS50EA finger oximeter recording at 60 Hz. The RGB videos were recorded with $1920 \times 1080$ pixels at 90 frames per second (fps) by TheImagingSource DFK 33UX290 camera. Videos were losslessly compressed with H.264 encoding using a constant rate factor of 0 to retain all raw video data and avoid damaging the optical pulse signal.

\paragraph{DDPM Dataset.} A total of 86 sets of recordings were collected, with each set consisting of nearly 11 minutes of video, for the {\it Deception Detection and Physiological Monitoring (DDPM)} dataset. During the recording, a paid actress conducted an interview consisting of 24 questions. Each subject was instructed beforehand to answer particular questions truthfully or deceptively. Subjects were free to complete the interview without constraints on motion, facial expressions, and talking, which accurately represents scenarios for unmasked pulse detection in the wild. The act of deception introduced variability in the pulse rate. {\bf Such variability is rarely observed in rPPG datasets, and thus overall, the dataset's size and collection setting make it unique}. The DDPM dataset is published along with this paper.

\paragraph{DDPM-Mask Dataset.} We augment the DDPM dataset with synthetic face masks by calculating a subset of face landmarks to define the occluded face region to create {\it DDPM-Mask} corpus. We use the same set of landmarks selected in \cite{NIST2020} to define a wide, medium coverage mask. For calculating the landmarks we use two landmarkers: the OpenFace (OF) toolkit \cite{Baltrusaitis2018} and Bulat \etal's 2D landmarker \cite{Bulat2017}. Along with a set of black masks, we also added patterned masks by randomly selecting images from the Describable Textures Dataset (DSD) \cite{Cimpoi14} and overlaying the image onto the 2D mask. With head rotation and translation, the pattern must be transformed to cover the same portions of the masked region. We first resized the pattern to the same dimensions as the input frames to the 3DCNN of $64 \times 64$ pixels. Then we randomly translated the pattern image such that the face landmarks for the first frame of the sequence were still within the pattern. Using these landmark points as anchors on the pattern image, we estimated the similarity transformation (rotation, translation, and scaling) from the anchor landmarks to the face landmarks in every following frame of the sequence, then applied the transformations on the pattern image before adding the masked region to the face frames. The second column of Fig. \ref{fig:main_figure} illustrates a patterned synthetic mask added to the DDPM dataset over a sequence of frames.

\paragraph{MPM Dataset.} We collected a new {\it Masked Physiological Monitoring (MPM)} video dataset for remote physiological monitoring of {\it masked} subjects, not participating in collection of the DDPM set. A plexiglass screen was placed between acquisition personnel and the subject to reduce COVID-19 transmission risk. Subjects were asked to bring 3 different face masks to increase the variability in color, texture, and shape.

We captured videos for 61 subjects over 3 different sessions, where the participant wore a different mask in each recording. We divided each session into three different tasks: (a) natural conversation with free head movement, (b) directed head movement, and (c) frontal view without head movement. The natural conversational task consisted of sustained interaction with an acquisition worker for 2 minutes. The directed head movement task aimed to stress the pulse detection algorithms by adding non-frontal gaze and head motion. Subjects were directed to look at a total of 6 different targets for approximately 5 seconds each, resulting in a 30 second interval. The final task consisted of the subject maintaining frontal gaze and avoiding movement or talking for 30 seconds. Poorly lit videos were removed resulting in a total of 170 usable recordings, over 3 minutes in length per video, giving us nearly 9 hours of recorded data. The reliability of the ground truth physiological signals was improved by using two Contec CMS50EA oximeters placed on both index fingers. The MPM dataset is published along with this paper.

\section{Approach}

We model the pulse prediction task as a regression problem with the blood volume waveform from the oximeter as the target. This task differs from popular action recognition tasks in the resolution of the prediction: in action recognition, a single class is predicted for a sequence of images, whereas our task generates a real value for every image in a sequence. We deploy a 3DCNN architecture on cropped frame sequences from the original video to contain the face only, as shown by Fig. \ref{fig:main_figure}. The following sections describe the model architecture and pipeline used to prepare the videos and target waveforms. 

\subsection{3DCNN Architecture}
We select the 3DCNN as the spatiotemporal architecture to learn the relation between frame sequences and cardiac waveform, 
similar to the PhysNet-3DCNN \cite{Yu2019}, but with modified temporal dimensions of the kernels from a width of 3 to a width of 9 to capture longer time dependencies and help filter out high-frequency noise. 

The 3DCNN was selected for three reasons. Firstly, it is capable of producing a high-resolution blood volume pulse waveform, not only selected statistics such as heart rate. Second, the 3DCNN is an end-to-end model, capable of learning from the raw image sequences. Lastly, the remote pulse detection task benefits from the joint learning of spatiotemporal features, rather than separating the dimensions and learning them independently.

\subsection{Video Preprocessing} \label{sec:vid_preproc}
To make the rPPG task easier for the model, we cropped the face region from all frames. We used two face landmarkers to detect 68 facial landmarks as a basis for defining the bounding box. Face landmarkers were used rather than simpler face detection models due to the stability of the landmark locations between adjacent frames. Detectors with less emphasis on fine-grained facial features add jitter to the bounding boxes over time, which in turn adds noise to the rPPG signal. Additionally, the face landmarks gave us keypoints to approximate the shape and location of a synthetic mask. We used two different approaches to landmarking for our experiments (the OpenFace (OF) toolkit \cite{Baltrusaitis2018} and the 2D landmarker of Bulat \etal (AB) \cite{Bulat2017}) to investigate reliance of the results on the landmarker. To create a bounding box, we found the minimum and maximum $(x,y)$ locations of the landmarks and extended the crop horizontally by 5\% to ensure that the cheeks and jaw were present. The top and bottom were extended by 30\%  and 5\% of the bounding box height, respectively, to include the forehead and jaw. From the extended bounding box, we further extended the shorter of the two axes to the length of the other to form a square region.

Taking insight from a recent rPPG study on the effect of image resolution for CNNs \cite{Zhan2020}, we downsized the cropped region to $64\times64$ pixels with bicubic interpolation. During training and evaluation, the model is given clips of the video consisting of 135 frames (1.5 seconds). We selected this as the minimum length of time an entire heartbeat would occur, considering 45 beats per minute (bpm) as a lower bound for healthy subjects.

\subsection{Physiological Signal Preprocessing}
The oximeters recorded ground truth waveform values at 60 Hz, which differed from the native 90 fps of the videos. Since our task requires a waveform label for every frame, we upsampled the ground truth waveform with cubic interpolation to the video timestamps for both DDPM and MPM collected datasets.

\begin{figure}
    \centering
    \includegraphics[width=0.76\linewidth]{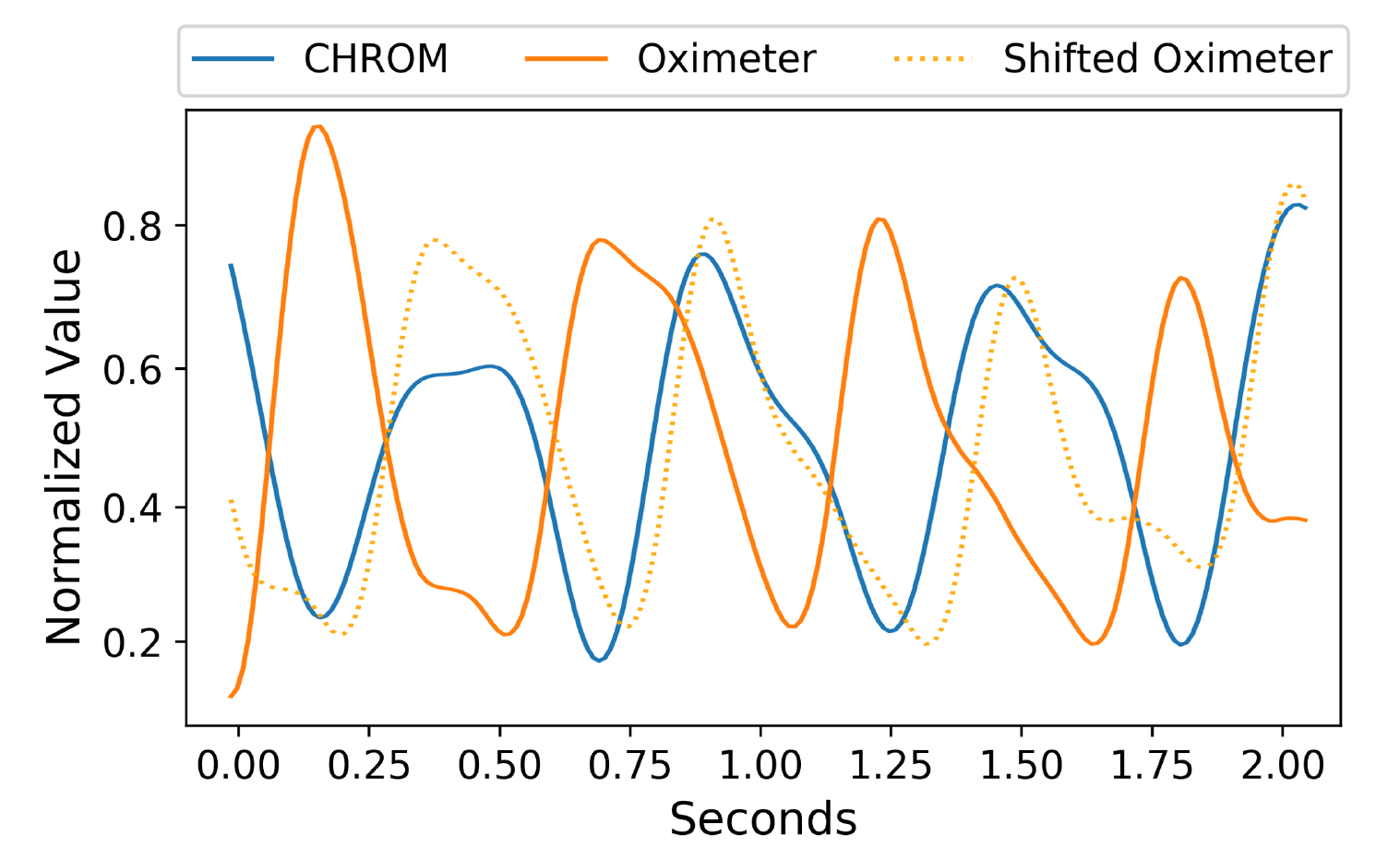}
    \caption{Ground truth waveform from the oximeter along with CHROM's prediction from the face region. Physiological and apparatus properties contribute to a phase shift between the two signals.}
    \label{fig:phase_shift}
\end{figure}

For training, phase differences between the pulse signal observed from the oximeter and the face present challenges. The relative phase of the blood volume pulse and the oximeter pulse is a function of both the subject's physiological structure and time lags from the acquisition apparatus. Zhan \etal \cite{Zhan2020} recently showed that a phase shift in the label when training a CNN dramatically reduces performance. To mitigate this issue, we applied the CHROM pulse detector \cite{DeHaan2013}, which is known to give reliable performance, to extract a reference waveform from the face to estimate the offset, and corrected the phase for the oximeter waveform as shown in Fig. \ref{fig:phase_shift}.

Given the finger and face waveforms, we calculated the cross-correlation between these signals with a sliding window of 10 seconds. A sliding window was used rather than the entire signals to allow for detrending \cite{Tarvainen2002} and normalization to minimize low-frequency differences between the two signals. Next, all windows were summed, and the location with a maximum sum within 1 second of lag was selected as the relative shift. All phase delays in the DDPM training set were found to be less than 0.4 seconds. For stable training of the 3DCNN we scaled the target waveform within each clip to real values in [0,1]. 

\subsection{Video Augmentation}
We augment the input data by horizontal flipping with 50\% probability, adding random illumination changes with mean of zero and standard deviation of 10 when operating on 8-bit grayscale images, and adding pixel-wise Gaussian noise $\sim{\cal N}(0,4)$. The image values are subsequently scaled to floating point values between 0 and 1. We augment every frame within each video clip in the same manner.

\subsection{Optimization and Training}
\begin{figure}
    \centering
    \includegraphics[width=\linewidth]{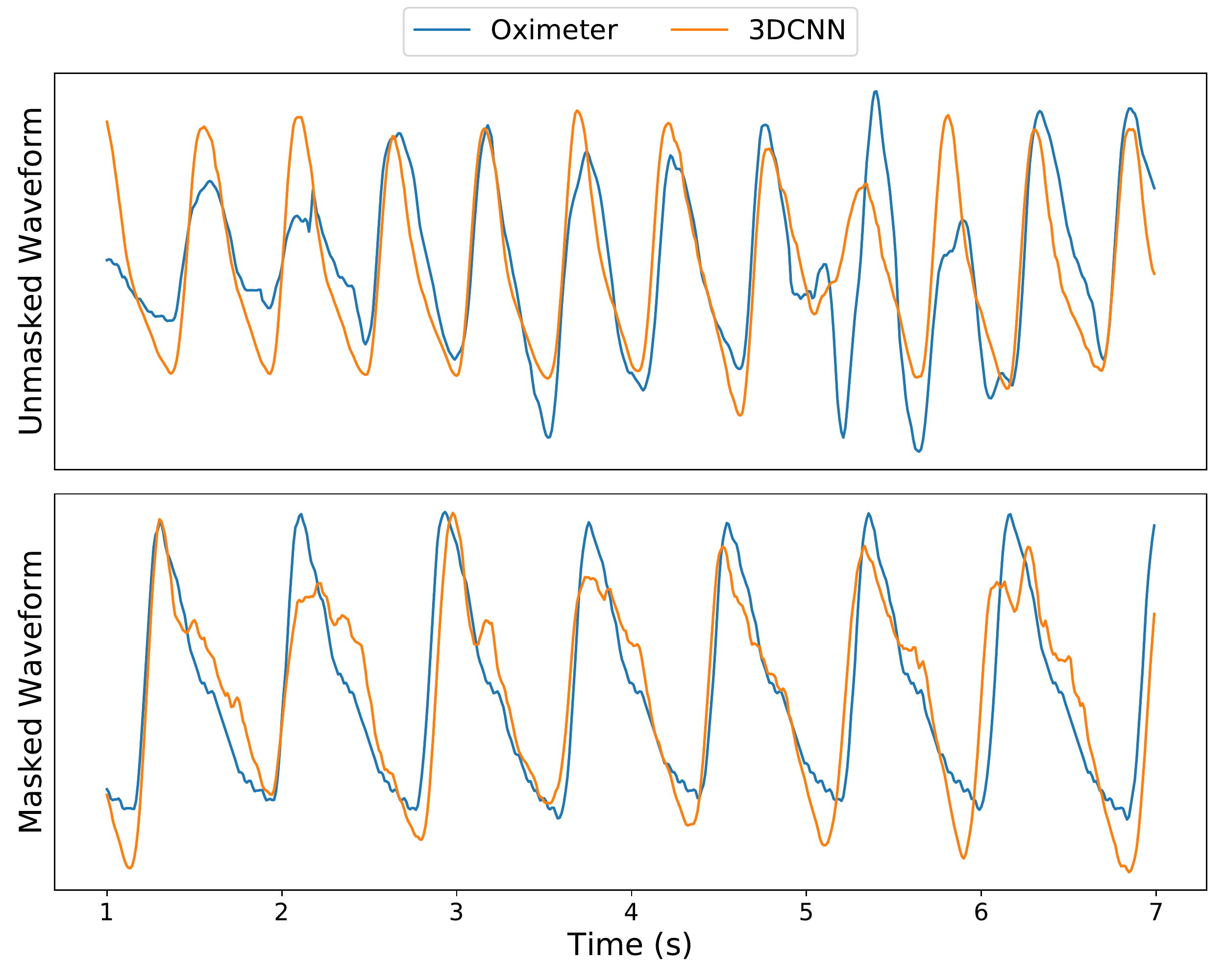}
    \caption{Ground truth and predicted waveforms for a short time segment on the unmasked (top) and masked (bottom) datasets using the same 3DCNN model trained on subjects without face occlusions.}
    \label{fig:unmasked_waves}
\end{figure}

We optimize the 3DCNN for the temporal regression problem by minimizing the negative Pearson correlation between waveforms, each of the length of 135 frames. We apply the Adam optimizer without weight decay, with a learning rate of $\alpha=0.0001$, and parameter values of $\beta_1=0.99$ and $\beta_2=0.999$. We apply dropout during training with 75\% probability, since 3DCNNs are prone to overfitting. Example ground-truth and predicted waveforms are shown in Fig. \ref{fig:unmasked_waves}. Visual inspection suggests that the trained 3DCNN model performs very well on both masked and unmasked faces.
 
\subsection{Overlap Adding}
The model is given short video clips and predicts a waveform value for every frame. For videos longer than the clip length (135 frames), it is necessary to perform predictions in sliding window fashion over the full video. Similar to \cite{DeHaan2013}, we use a stride of half the clip length to slide across the full video. The windowed outputs are first standardized, then a Hann function is applied to mitigate edge effects from convolution by weighting the window's center more than the extremes. Finally all overlapped outputs are summed to give a final waveform, as shown in Fig. \ref{fig:main_figure} (right).
\section{Experimental Protocol}

\subsection{Experimental Scenarios}
\label{sec:scenarios}

Our experiments attempt to understand how face masks adversely affect remote pulse detection performance, and whether adding synthetically-generated masks to face videos during training helps improve performance in the presence of real face masks. To give a complete evaluation, we evaluate all models on both the masked (MPM) and unmasked (DDPM) datasets, and with two different face landmarkers, in the following {\bf four scenarios}:

\begin{itemize}
\setlength{\itemsep}{0pt}
    \item[(s1)] training / tuning all methods on {\bf unmasked} face videos (train/validation partition of DDPM), and testing also on {\bf unmasked} face videos (test partition of DDPM),
    \item[(s2)] training / tuning all methods on face videos with {\bf synthetically added masks} (DDPM-Mask dataset), and testing on {\bf unmasked} subject-disjoint face videos (test partition of DDPM),
    \item[(s3)] training / tuning all methods on {\bf unmasked} face videos (train/validation partition of DDPM), and testing on {\bf masked} face videos (MPM dataset),
    \item[(s4)] training / tuning all methods on face videos with {\bf synthetically added masks} (DDPM-Mask dataset), and testing on {\bf masked} face videos (MPM dataset).
\end{itemize}

\subsection{Dataset Partitions}
The unmasked dataset (DDPM) was divided into three subject-disjoint partitions: 64 subjects used for training, another 11 subjects used for validation, and the remaining 11 subjects used for testing. The splits were crafted with stratified random sampling across race, gender, and age, in order of importance in the cases that equal splits were not possible. By setting a portion of the unmasked data (DDPM) aside for testing, we can effectively examine the change in performance when evaluating on the entire masked (MPM) dataset of 61 subjects.

\subsection{Compared Methods}
We selected several previous state-of-the-art algorithms to evaluate the efficacy of our approach. To the authors' knowledge, few recent methods are open source, and implementing such techniques from scratch may be difficult from the papers alone. We requested code from members of the pulse detection research community and were unsuccessful in getting source codes. All hand-crafted methods evaluated in the paper, including chrominance-based (CHROM) \cite{DeHaan2013} and plane-orthogonal-to-skin (POS) \cite{Wang2017}, were reimplemented by us with minor help from components of Heusch \etal \cite{Heusch2017}.

Two algorithms employing blind-source separation of the color channels through independent component analysis (ICA) \cite{Poh2010, Poh2011} were also tested, due to their initial popularity in the field. For simplicity, we refer to the ICA approach presented in \cite{Poh2010} as POH10, and refer to the improved ICA approach with detrending \cite{Poh2011} as POH11. Both of the ICA approaches perform spatial averaging on the cropped facial region after applying a face detector. We apply OpenFace and use the landmarks to define the region of interest in the same protocol presented in section \ref{sec:vid_preproc}.

Unfortunately, code or the weights could not be acquired for recent deep learning-based approaches \cite{Chen2018, Niu2020}. We use the previously described 3DCNN as an examplar for the deep learning approaches. Given the output waveforms from the 4 handcrafted approaches, in addition to the proposed 3DCNN trained on DDPM with OpenFace (OF) and Bulat \etal (AB) landmarking approaches (denoted in the results as 3DCNN OF and 3DCNN AB), DDPM-Mask with black synthetic masks from both landmarkers (3DCNN OF+B and 3DCNN AB+B), and DDPM-Mask with patterned synthetic masksfrom both landmarkers (3DCNN OF+P and 3DCNN AB+P), we evaluate each method in the exact same manner for a fair comparison.

The efforts described above related to acquisition, re-implementation of various rPPG methods, and their evaluation may be regarded as the best comparison to SOTA rPPG methods one can do today.

\subsection{Evaluation Metrics}
We evaluated the model performance in both the {\bf temporal} (associated with the waveform shape) and {\bf frequency} (associated with the heart beat rate) domains. Nearly all past works evaluate the heart rate prediction which may be not adequate for justifying deployment on many vital signs tasks. In recent works, blood oxygenation and blood pressure have been predicted from high quality photoplethysmograms \cite{Guazzi2015, Martinez2018}, motivating evaluation of the pulse waveforms in the temporal domain as well.

Evaluating the predictions in the {\bf temporal domain} was accomplished by examining three-second windows of the waveforms. We calculated the Pearson correlation between the normalized ground truth and predicted waveforms with a stride of 1 frame between windows ($r_t$). The short time windows were selected due to the difference in the amplitude of the signals over longer periods, especially if they were affected by low frequency components causing trends. We only performed the temporal analysis for the unmasked (DDPM) dataset, since we were able to extract accurate waveforms with CHROM to correct the phase differences. For the masked (MPM) dataset, existing algorithms performed poorly, so we were unable to find a phase shift to evaluate fine-grained waveform differences.

The pulse detection performance in the {\bf frequency domain} is analyzed by calculating the error between heart rates, defined by the dominant frequency in the waveform for short time periods. A recent work showed that the time window size for predicting heart rate can significantly affect error estimation \cite{Mironenko2020}. Since the time window used to predict heart rate within the oximeter is unknown, we calculate the ground truth heart rate frequencies from the oximeter's waveforms. Specifically, we use a sliding window of length 30 seconds and apply a Hamming window prior to converting the signal to the frequency domain with the Fast Fourier Transform (FFT). The frequency index of the maximum spectral peak between $0.\overline{66}$ Hz and 3 Hz (40 bpm to 180 bpm) is selected as the heart rate. A five-second moving average filter is then applied to the resultant heart rate signal to smooth noisy regions containing finger movement. To compare the heart beat estimates, we used standard metrics from the rPPG literature, such as mean error (ME), mean absolute error (MAE), root mean squared error (RMSE), and Pearson correlation coefficient ($r_f$) between heart rate predictions.

Since two oximeters are present in the masked dataset, we perform the same procedure over both waveforms and average the heart rate value at each time step. In a very small number of cases, the noise from hand movement gave different heart rate values from the oximeters. To remove these portions, if the heart rates differed by more than 10 beats per minute, the calculated heart rate closest to the average of the original heart rate estimates from the oximeters was selected. The resultant signals were smoothed with a three-second moving average filter to avoid spurious jumps in the heart rate. Ground truth heart rate values were verified by manually finding peak-to-peak distances in a subset of the waveforms, and these calculations were found to be more robust than applying standard peak detectors. 

\section{Results}

\begin{table}\small
    \begin{center}
    \caption{Pulse rate estimation comparison when the methods are tested on videos without face masks (scenarios s1 and s2). ``B'' and ``P'' denote black mask and patterned synthetic masks added to the training data, respectively.}
    \label{tab:unmasked_tests}
    \begin{tabular}{ccccccc}
        \toprule
        Method & \begin{tabular}{@{}c@{}}ME \\ (bpm)\end{tabular} & \begin{tabular}{@{}c@{}}MAE \\ (bpm)\end{tabular} & \begin{tabular}{@{}c@{}}RMSE \\ (bpm)\end{tabular} & \begin{tabular}{@{}c@{}}$r_f$\\ (bpm)\end{tabular} & 
        \begin{tabular}{@{}c@{}}$r_t$\\ (wave)\end{tabular} \\
        \midrule\\[-2.5ex]
        CHROM \cite{DeHaan2013} & -0.26 & 3.48 & 10.37 & 0.93 & 0.60 \\
        POS \cite{Wang2017}  & \textbf{0.11} & 3.16 & 11.19 & 0.92 & -0.44 \\
        POH10 \cite{Poh2010} & 18.54 & 20.56 & 33.10 & 0.56 & -0.12 \\
        POH11 \cite{Poh2011} & 10.47 & 14.30 & 28.86 & 0.54 & 0.10 \\
        3DCNN OF & -1.18 & \textbf{1.96} & \textbf{6.99} & \textbf{0.97} & 0.68 \\
        3DCNN AB & -1.25 & \textbf{1.96} & 7.17 & \textbf{0.97} & \textbf{0.70} \\
        \hline \\[-2.2ex]
        3DCNN OF+B & -1.18 & 2.06 & 7.29 & \textbf{0.97} & 0.68 \\
        3DCNN AB+B & -1.16 & 2.03 & 7.28 & \textbf{0.97} & 0.69  \\
        3DCNN OF+P & -1.27 & 2.00 & 7.29 & \textbf{0.97} & 0.69 \\
        3DCNN AB+P & -0.81 & 2.30 & 7.76 & 0.96 & 0.67 \\
        \bottomrule
    \end{tabular}
    \end{center}
\end{table}

\begin{table}\small
    \begin{center}
    \caption{Same as in Tab. \ref{tab:unmasked_tests} except that the methods are tested on videos with face masks (scenarios s3 and s4).}
    \label{tab:masked_tests}
    \begin{tabular}{ccccccc}
        \toprule
        Method & \begin{tabular}{@{}c@{}}ME \\ (bpm)\end{tabular} & \begin{tabular}{@{}c@{}}MAE \\ (bpm)\end{tabular} & \begin{tabular}{@{}c@{}}RMSE \\ (bpm)\end{tabular} & \begin{tabular}{@{}c@{}}$r_f$ \\ (bpm)\end{tabular} \\
        \midrule\\[-2.5ex]
        CHROM \cite{DeHaan2013} & 3.05 & 12.59 & 16.29 & 0.02 \\
        POS \cite{Wang2017} & 15.80 & 18.84 & 26.56 & 0.13 \\
        POH10 \cite{Poh2010} & 25.11 & 26.74 & 32.76 & -0.02 \\
        POH11 \cite{Poh2011} & 38.29 & 38.31 & 40.77 & -0.07 \\
        3DCNN OF & -1.45 & 3.57 & 9.38 & 0.79 \\
        3DCNN AB & -2.15 & 3.87 & 10.49 & 0.76 \\
        \hline \\[-2.2ex]
        3DCNN OF+B  & -1.64 & 3.81 & 9.70 & 0.78 \\
        3DCNN AB+B  & -1.80 & 3.71 & 9.68 & 0.78 \\
        3DCNN OF+P  & -1.88 & 3.75 & 9.47 & 0.79 \\
        3DCNN AB+P  & \textbf{-0.67} & \textbf{3.36} & \textbf{8.47} & \textbf{0.81} \\
        \bottomrule
    \end{tabular}
    \end{center}
\end{table}

This section provides results and discussion for all four experimental scenarios described in Sec. \ref{sec:scenarios}.

\paragraph{Scenario s1 (baseline): training and testing on {\it unmasked} face videos.} Performance for unmasked participants (DDPM) is shown in the {\bf top portion of Table \ref{tab:unmasked_tests}}. The two chrominance-based methods achieve lower mean error rates, showing that they are well calibrated for predicting heart rate and don't exhibit bias. Both ICA-based methods give worse performance than the chrominance and 3DCNN approaches on every metric. The 3DCNN model contains slightly worse mean error rates from bias than the chrominance models, but performs remarkably well in terms of MAE and RMSE. The choice of landmarker does not appear to affect performance significantly.

\paragraph{Scenario s2: training on face videos {\it with synthetic masks}, testing on {\it unmasked} face videos.} The results for models trained on synthetically masked participants (DDPM-Mask) are shown in the {\bf bottom portion of Table \ref{tab:unmasked_tests}}. Error discrepancies between the black and patterned masks are negligible, considering they perform better on different metrics. We find that models trained with synthetic masks generally perform worse than the model trained to use the entire facial region, since the signal to noise ratio is decreased.

\paragraph{Scenario s3: training on {\it unmasked} face videos, testing on videos of faces wearing {\it real masks}.} Performance of the handcrafted methods and 3DCNN model trained on DDPM and evaluated on masked subjects (MPM) are shown in the {\bf upper portion Table \ref{tab:masked_tests}}. In general, performance degrades substantially compared to maskless. The best MAE among the handcrafted methods is given by CHROM, with over 13 bpm -- more than 3 times worse than on DDPM. The 3DCNN model gives significantly better performance than the chrominance and ICA approaches, although performance is generally worse. Fortunately, correlation between heart rate predictions and ground truth remains strong for the 3DCNN, with $r_f = 0.79$. For general purposes, the increase in error is likely not large enough to change an assessment of one's state of health, but improving performance to the unmasked baseline is desirable. The performance decrease indicates that face occlusions cause difficulties for all analysed approaches.

\paragraph{Scenario s4: training on face videos {\it with synthetic masks}, testing on videos of faces wearing {\it real masks}.} The {\bf lower portion of Table \ref{tab:masked_tests}} shows the performance of the models trained on black (3DCNN OF+B and 3DCNN AB+B) and patterned (3DCNN OF+P and 3DCNN AB+P) synthetic masks. We don't find an improvement with black synthetic masks, however, we see the best-performing approach is trained with patterned synthetic masks and Bulat \etal's landmarker, giving the lowest ME, MAE, and RMSE, along with the highest correlation between heart rates. Interestingly, the patterned synthetic masks with the OpenFace landmarker do not help the model, showing that the choice of landmarker is important.

\section{Discussion}

\begin{figure}
\centering
\begin{subfigure}[b]{.5\linewidth}
\centering
    \includegraphics[width=0.8\linewidth]{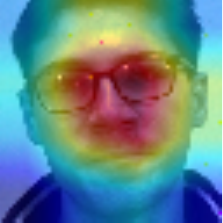}
    \caption{3DCNN OF}
    \label{fig:ofddpm}
\end{subfigure}%
\begin{subfigure}[b]{.5\linewidth}
\centering
    \includegraphics[width=0.8\linewidth]{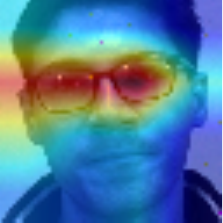}
    \caption{3DCNN AB+P}
    \label{fig:abpatddpm}
\end{subfigure}
\begin{subfigure}[b]{.5\linewidth}
\centering
    \includegraphics[width=0.8\linewidth]{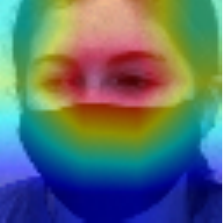}
    \caption{3DCNN OF}
    \label{fig:ofmpm}
\end{subfigure}
\begin{subfigure}[b]{.5\linewidth}
\centering
    \includegraphics[width=0.8\linewidth]{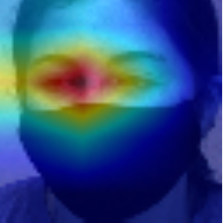}
    \caption{3DCNN AB+P}
    \label{fig:abpatmpm}
\end{subfigure}
\caption{Grad-CAM heatmaps for the top performing networks on DDPM (3DCNN OF) and MPM (3DCNN AB). The top and bottom rows show samples from the DDPM and MPM datasets, respectively. The attended regions for the model trained on DDPM covers a larger portion of the face than the the model trained on DDPM-Mask, which was guided to focus on the periocular and forehead regions. Images are scaled for viewing purposes.}
\label{fig:gradcam}
\end{figure}

\paragraph{Visual Explanations.} We apply Grad-CAM \cite{Selvaraju2017} to visually explain the performance differences between the 3DCNNs. Since Grad-CAM is traditionally used for single images, we collected the pixel-wise sum over all images in a clip followed by normalization for image viewing. We then overlay the heatmap over the middle frame of the sequence. Figure \ref{fig:gradcam} shows the attended spatial regions in the eighth convolutional layer for 3DCNN OF and 3DCNN AB+P, the best performing models for unmasked and masked subjects, respectively. The heatmaps clearly show that 3DCNN OF attends to the center of the face region, even when partially occluded by a face mask, while 3DCNN AB+P has learned to attend to the periocular region and forehead, since the synthetic masks during training occluded the lower face. This helps explain why the performance for 3DCNN AB+P is impressive on masked faces and worse on unmasked faces, where the full face is not utilized.

\begin{figure}
\centering
\begin{subfigure}{.5\linewidth}
    \centering
    \includegraphics[width=0.9\linewidth]{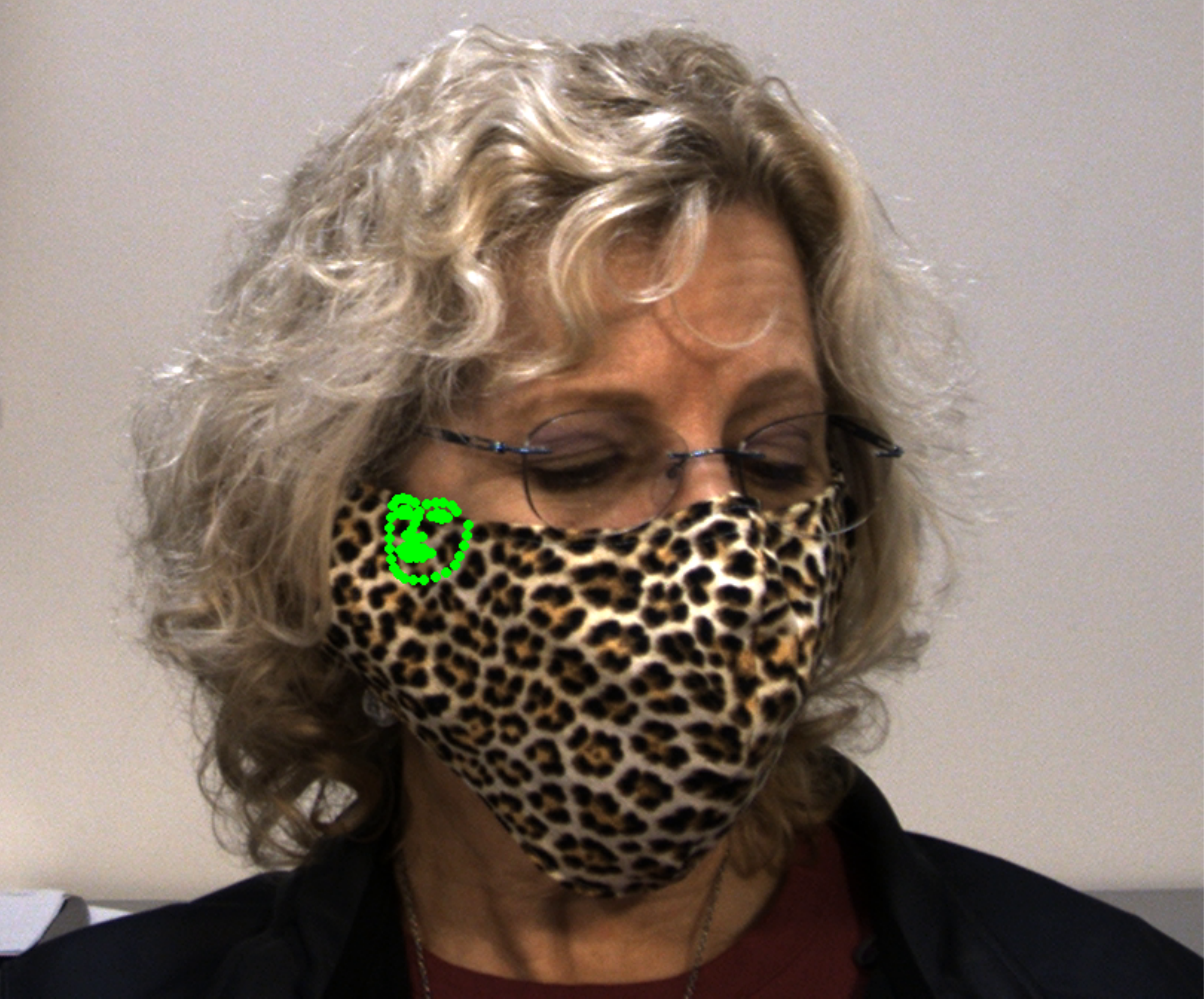}
    \label{fig:bad_landmarks1}
\end{subfigure}%
\begin{subfigure}{.5\linewidth}
  \centering
    \includegraphics[width=0.9\linewidth]{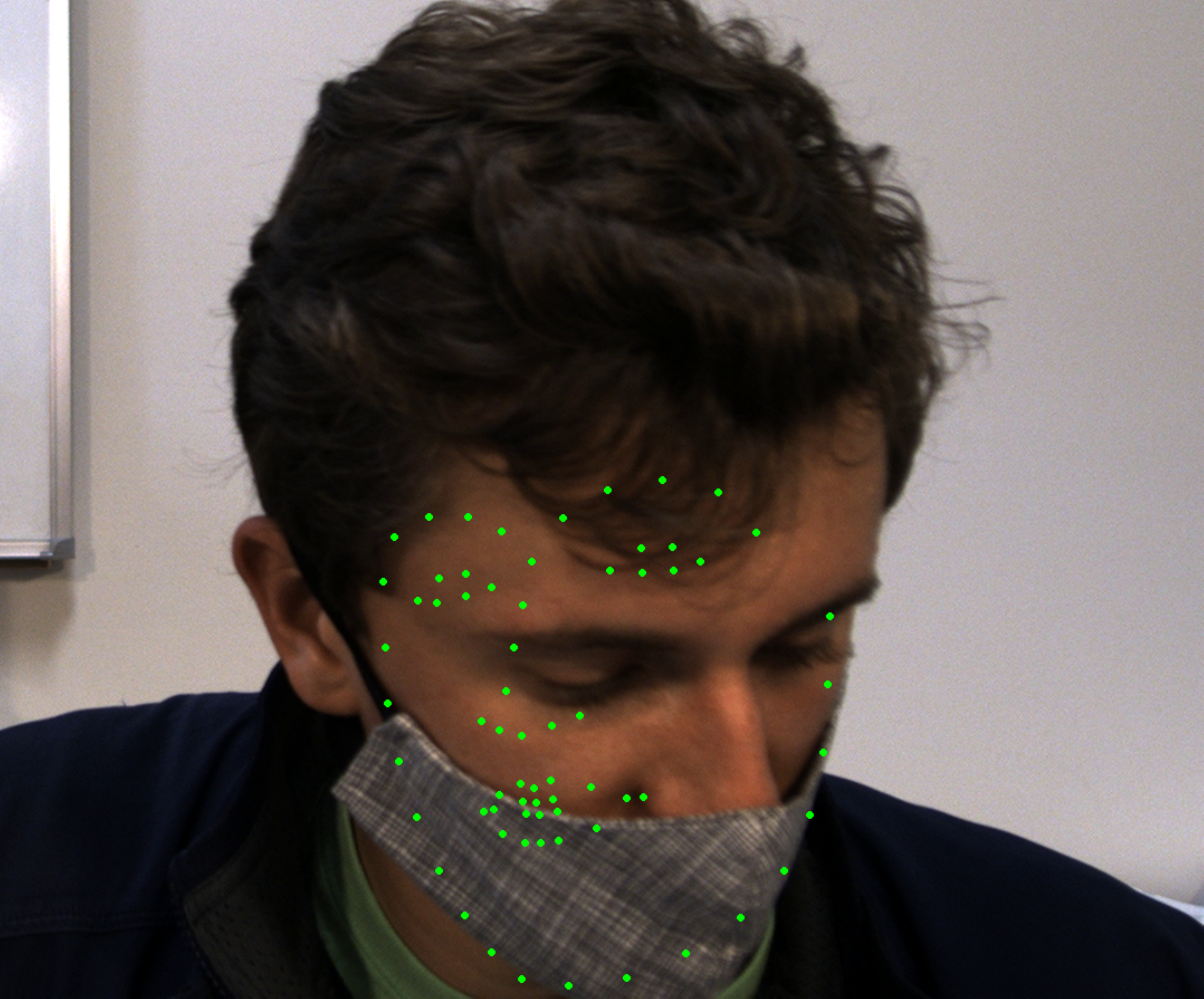}
    \label{fig:bad_landmarks2}
\end{subfigure}
\caption{Face masks present difficulties to face detection and landmarking algorithms, as shown by the errors when using OpenFace on heavily occluded faces.}
\label{fig:bad_landmarks}
\end{figure}

\paragraph{Importance of Landmarker.} Our experimental results show the importance of selecting an accurate landmarker when generating the synthetically masked videos. When using the OF landmarker, the models trained on black and patterned synthetic masks performed slightly worse than the model trained on unmasked faces. Figure \ref{fig:bad_landmarks} shows erroneous face landmarks produced by the OF method when subject gazed away from the cameras, which occurs frequently in the DDPM dataset due to the interview scenario. The opposite result occurs when using the AB landmarker, where training on synthetic masks improves performance and even gives the best performance of all approaches.

\section{Conclusions}
In this paper, we present a new large-scale physiological monitoring dataset of high resolution RGB video and oximeter recordings to give the first evaluation of remote pulse estimators on masked subjects.

In {\bf answering the research questions} posed in the introduction, we find: (re: Q1) accurate pulse estimation is possible when subjects are wearing face masks, but the performance is slightly worse, (re: Q2) training with synthetically generated mask videos improves performance {\it when using robust face landmarkers}, and (re: Q3) face landmarkers and skin detectors robust to heavy face occlusion should be deployed in the early phases of the pulse detection algorithms to define reliable regions of interest. Several previous state-of-the-art pulse estimators built for unoccluded face video are found to perform substantially worse on masked subjects, while a 3DCNN exhibits a moderate drop in performance. We find training the model with patterned synthetic masks created with accurate face landmarkers is sufficient to increase the robustness of pulse detection in the presence of  masks and close the gap in performance.

\section*{Acknowledgements}
We would like to thank Marybeth Saunders for conducting the interviews during DDPM data collection. This research was sponsored by the Securiport Global Innovation Cell, a division of Securiport LLC.  Commercial equipment is identified in this work in order to adequately specify or describe the subject matter. In no case does such identification imply recommendation or endorsement by Securiport LLC, nor does it imply that the equipment identified is necessarily the best available for this purpose.  The opinions, findings, and conclusions or recommendations expressed in this publication are those of the authors and do not necessarily reflect the views of our sponsors. 

{\small
\bibliographystyle{ieee_fullname}
\bibliography{egbib}
}


\end{document}